\documentclass{article}
\usepackage{spconf,amsmath,graphicx}

\usepackage{booktabs}

\title{Deep Learning of Segment-Level Feature Representation for Speech Emotion Recognition in Conversations}
\name{Jiachen Luo$^1$, Huy Phan$^{2*}$\thanks{$^{*}$The work was done when H. Phan was at Centre for Digital Music, Queen Mary University of London, UK and prior to joining Amazon.}, Joshua Reiss$^1$\thanks{Thanks to the China Scholarship Council and Queen Mary University of London for funding.}}
\address{$^1$Centre for Digital Music, Queen Mary University of London, UK\\$^2$Amazon Alexa, Cambridge, MA, USA}
\begin{document}
\ninept
\maketitle
\begin{abstract}
Accurately detecting emotions in conversation is a necessary yet challenging task due to the complexity of emotions and dynamics in dialogues. The emotional state of a speaker can be influenced by many different factors, such as interlocutor stimulus, dialogue scene, and topic. In this work, we propose a conversational speech emotion recognition method to deal with capturing attentive contextual dependency and speaker-sensitive interactions. First, we use a pretrained VGGish model to extract segment-based audio representation in individual utterances. Second, an attentive bi-directional gated recurrent unit (\emph{GRU}) models contextual-sensitive information and explores intra- and inter-speaker dependencies jointly in a dynamic manner. The experiments conducted on the standard conversational dataset MELD demonstrate the effectiveness of the proposed method when compared against state-of the-art methods.
\end{abstract}


\section{Introduction}
\label{sec:intro}

Automatic recognition of human emotions has widespread applications in areas such as dialogue generation, social media analysis and human computer interaction [1]. Speech is the main communication medium in which people can clearly and intuitively feel emotional changes. Unlike vanilla emotion recognition of sentences/utterances, emotion recognition in conversation (ERC) ideally relies on mining human emotions from conversations or dialogues having two or more interlocutors and requires context modeling of the individual utterances, and requires context modeling of the individual utterances [2]. How to capture such information from speech signal is a challenging task.

In this work, we focus on speech signals in interactive conversation. Speech signals naturally can carry the emotional characteristics of the speaker. Conventionally, conversational emotion recognition usually requires a strong ability to model context-sensitive attributes, select crucial information, and capture speaker-sensitive dependencies [3]. Among all the factors, speaker information is important for tracking the emotional characteristics of conversations, especially intra- and inter-speaker dependencies.

In interactive conversations, these factors lead to diverse emotional dynamics. Fig. 1 presents some examples demonstrating such patterns from the Multi-modal EmotionLines Dataset (MELD) [4]. Conversation (a) depicts the presence of emotional inertia which speakers influence on themselves. The character Ross maintains a neutral emotional state by not being influenced by the other speaker. On the other hand, conversation (b) refers to inter-speaker dependencies that counterparts induce in a speaker. ``I have to buy a new one?'' shows negative attitude. With a particular voice shade of anger, it can affect the feeling of the addressee/listener. ``How’d you get to that?'' emphasizes to the listener that Joey is affected by the feeling of speaker Chandler responses.

\begin{figure}[t!]

\begin{minipage}[b]{1.0\linewidth}
  \centering
  \centerline{\includegraphics[width=8.5cm]{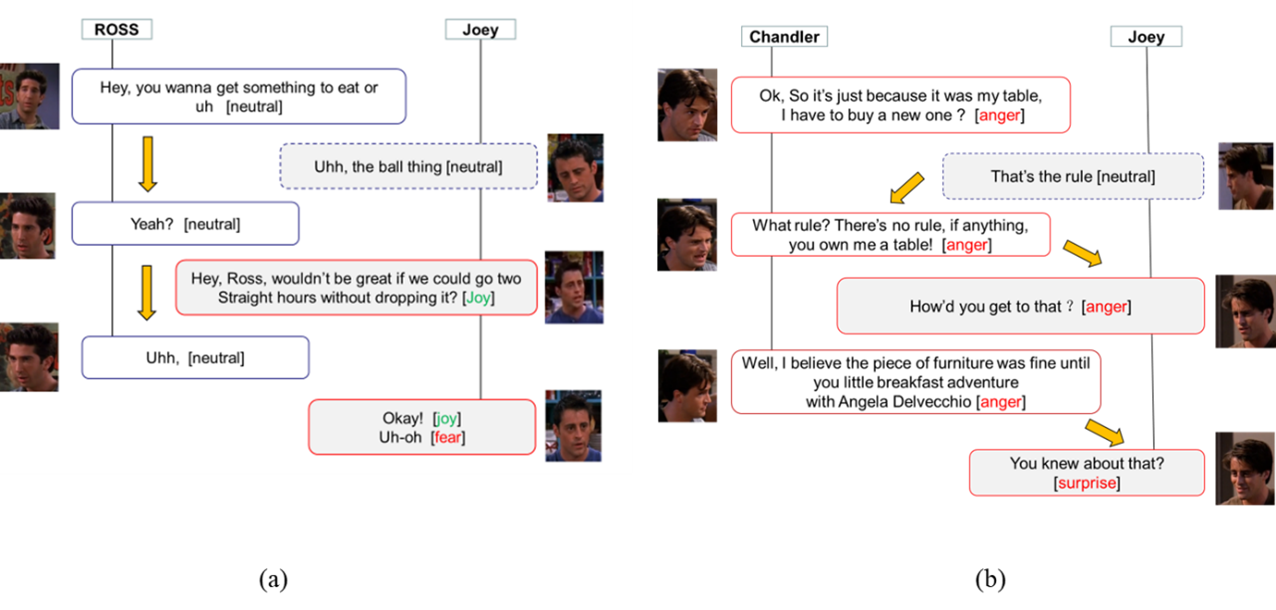}}
\end{minipage}
\caption{\small Emotion dynamic of speakers in a dialogue in comparison.}
\end{figure}

To model such conversations, an architecture would need to deal with these challenges: how to capture self- and inter-speaker dependencies to govern emotional dynamics, and how to interpret latent emotions from its contextual information in the conversation flows. What’s more, the raw emotion can be enhanced, weakened, or reversed based on the contextual information from neighboring utterances [5]. For utterance-level speech emotion recognition, an underlying issue is a loss of dynamic temporal information and short-term emotion dynamics  by compressing speech into utterance-level features [6]. However, little progress has been made in analyzing the emotion estimation among segment-based feature representation in individual utterances, context-sensitive information and speaker influences in conversations. Devamanyu et al. used text modality features to model the contextual information into self- and inter-speaker emotional influences in the ERC task [7]. 

In this paper, we present an approach which can enable the co-evolution of the local and global contextual information among segments in utterances and accommodate the self-influence, intra- and inter-speaker state in the emotion-aware spoken dialog system. We first use a pre-trained VGGish model to extract segment-level audio representation in an utterance. Next, a statistical strategy determines emotional dynamics of utterance-level information in speech. To dynamically integrate contextual dependency and speaker-sensitive interactions, we employ bi-directional \emph{GRU} to model such relations. Overall, our contributions are summarized as follows:

$\bullet$ A segment-level feature extraction strategy was able to empower a dialogue system with fine-grained temporal emotional representation. 

$\bullet$ We utilized the bi-directional \emph{GRU} layer to capture context-sensitive information, intra- and inter-speaker dependencies on conversational emotion recognition, in combination with attention mechanism to highlight the important global contextual utterances. 

$\bullet$ Our proposed approach was shown to be superior to state-of-the-art methods for conversational emotion recognition.

\section{Existing Literature}
\label{sec:format}

Global and local audio features of speech emotion recognition systems are typically classified into the following four categories: prosodic features, spectral features, voice quality features, and Teager Energy Operator-based features [8]. Traditionally, a number of spectral features are generally depicted using one of the cepstrum-based representations available. Commonly, Mel-frequency cepstral coefficients (MFCC) or Mel-scale spectrograms were used, and in some studies, formants, and other information were utilized as well [8]. Suraj et al. demonstrated the effectiveness of convolutional neural networks in emotion classification with MFCCs [5]. Besides, the direct use of Mel-scale spectrograms for ERC was proved successful as well [9]. In this work, we use a pretrained model to extract high-level acoustic features from low-level Mel-scale spectrograms for emotion recognition.  

ERC requires deep understanding of human interactions in conversations [10$\sim$13]. Some of the important works attributes emotional dynamics to be interactive phenomena [13,16], rather than being within-person. We utilize this trait in the design of our model that incorporates inter-speaker dynamic in a conversation. Since conversations have a natural temporal nature, context also play a crucial role in emotion analysis [4,17]. Poria et al. employed a bi-directional LSTM to capture temporal context information from surrounding utterances of the same speaker to infer emotions [4]. However, there is no provision to model context and speaker interactive influences. 

It should be noted that our method is different from previous strategies for ERC. We propose to capture this contextual-sensitive information via hierarchical recurrent networks. Additionally, our proposed approach adopts an interactive scheme that actively models intra- and inter-speaker emotional dynamics in conversations.

\section{Methodology}
\label{sec:typestyle}

Our proposed approach consists of three major modules: 1) pre-trained segment-level audio representation, 2) grouped parallel statistical operation, referred to as statistical unit (SU), and 3) a dialog-aware interaction module that aims to model the interactions in the dialogue and then makes the emotional state prediction.

\subsection{ Pre-Trained Audio Embeddings}
\label{ssec:subhead1}
On the audio end, an audio segment of 0.96s is converted into log Mel-spectrogram and encoded using a 128-dimensional feature vector extracted from the last fully connected layer of a pre-trained VGGish model. Especially, the shorter samples were zero-padded before transformation into log Mel-spectrogram. Non-overlapping segments are used during segmentation. The model was trained on AudioSet, consisting of 100 million YouTube videos [14]. For each utterance, a sequence of \emph{L} embedding vectors was produced, where \emph{L} represents the number of audio segments that the audio signal is partitioned into (see Fig. 2). 

\subsection{Statistical Unit}
\label{ssec:subhead2}
We use a statistical unit with three parallel one-dimensional statistics along the sequence direction to reduce the sequence of \emph{L} segment-wise embedding vectors and produce utterance-wise embedding vectors: average, max and min, as shown in Fig. 2. Finally, we concatenate them into one feature vector for utterance-wise representation.

\begin{figure}[t!]

\begin{minipage}[b]{1.0\linewidth}
  \centering
  \centerline{\includegraphics[width=8.5cm]{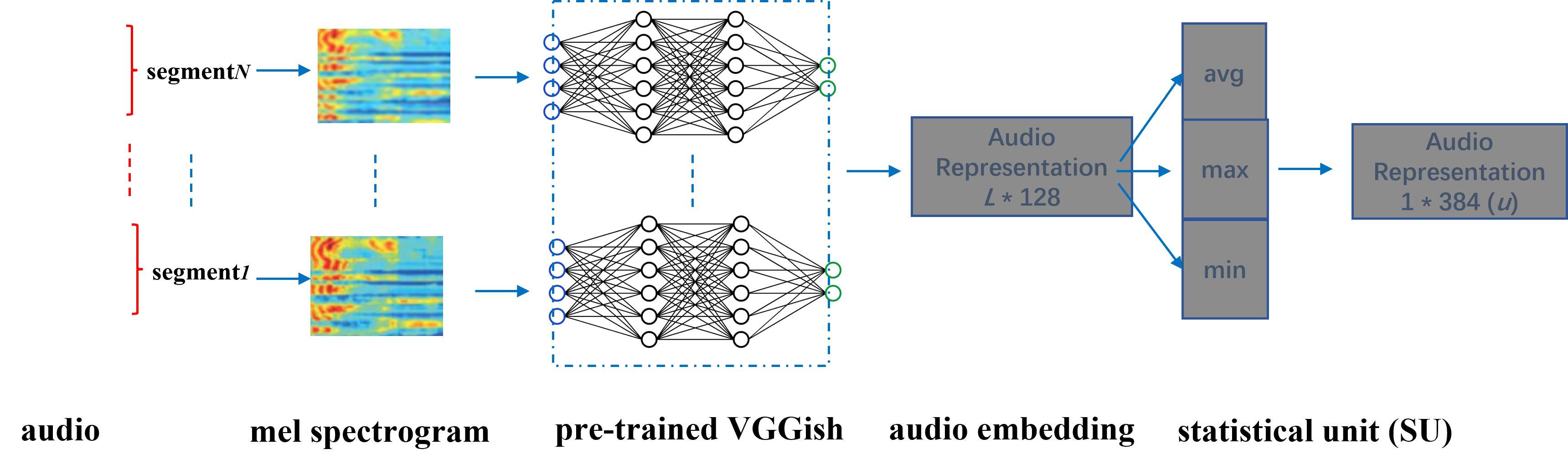}}
\end{minipage}
\caption{\small The aggregation of segment-level embedding extracted from a pre-trained VGGish model to produce utterance-wise embedding.}
\end{figure}

\subsection{Model}
\label{ssec:subhead3}
 The proposed model has four branches of bi-directional \emph{GRU} cells to capture the cumulative context, intra-speaker state, inter-speaker influence and emotion state of the participant (see Fig. 3).

\subsubsection{Attentive Contextual State}
\label{sssec:subsubhead1}
In conversational emotion recognition, to determine the emotional state of an utterance at timestamp $t$, the preceding utterances can be considered as its cumulative context. The context state stores and propagates overall utterance-level information along the sequence of the conversation flow. The contextual state $C_{t-1}$, intra-speaker state $S_{t-1}$ and inter-speaker state $I_{t-1}$ of the previous utterance, and audio representation \emph{u}$_t$ at timestamp $t$ are used to update the contextual information from $C_{t-1}$ to $C_t$ (see Fig. 3). The steps in the attentive contextual state update $C_t$ are described using the following formula and shown in Fig. 3.
\begin{align}
C_t = GRU_C(C_{t-1}, (S_{t-1}\oplus I_{t-1}\oplus u_t))
\end{align}
where $\oplus$ represents concatenation. At the time step $t=0$, the context state is randomly initialized.

In order to amplify the contribution of the context-rich information, we employ soft-attention from the history interactive context to combine long-context speaker interaction influences and conversational dependence [15]. We pool the attention vector $a_t$ from the surrounding context history $[C_1,C_2,\ldots, C_{t-1}]$ using soft-attention. This contextual attention vector $a_t$ can be computed as follows: 
\begin{align}
u_i &= \tanh(WC_i +b), \mbox{~~} 1 \le i \le t-1, \nonumber \\
\alpha_i &= \frac{\exp(u_i^\mathsf{T})}{\sum_{i=1}^{t-1} {\exp(u_i^\mathsf{T})}}, \nonumber \\
a_t &= \sum\nolimits_{i=1}^{t-1} {\alpha_i}{C_i}.
\end{align}

\begin{figure}[htb]

\begin{minipage}[b]{1.0\linewidth}
  \centering
  \centerline{\includegraphics[width=8.5cm]{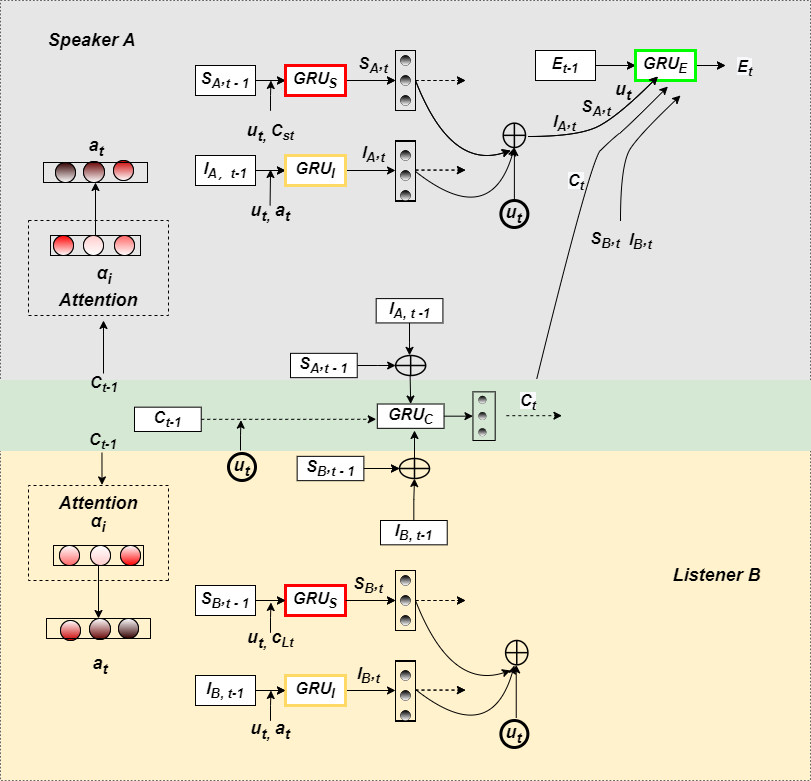}}
\end{minipage}
\caption{\small The overall architecture of our proposed model for emotion prediction in conversation.}
\end{figure}

\subsubsection{Self-Speaker State}
\label{sssec:subsubhead2}
The self-influence is conditioned on how the speakers tend to maintain emotions during the conversations. This state is also known as emotional inertia, as speakers may not always express explicitly their feeling or outlook through reactions. Concretely, self-influence only involves speaker himself/herself. 

Self-Influence Module consists of two $GRU_{S\lambda}$: $GRU_{SS}$ and $GRU_{SL}$, respectively. For For $\lambda \in \{S, L\}$, $S$ denotes the speaker and $L$ denotes the listener. $GRU_{S\lambda}$ attempts to memorize the emotional inertia of $P_{\lambda}$
which represents the emotional dependency of the person with their own previous states. A dialogue involves two parties, the speaker and others belonged to the listener. For time step $t$, the self-speaker state of the person $S_{\lambda t}$ is updated by the previous self-speaker state of the person $S_{\lambda(t-1)}$, and the cumulative contextual vector $C_{\lambda t}$, and the utterance $u_{\lambda t}$. At time step $t$, the self-speaker state $S_{\lambda t}$ can be computed as:
\begin{align}
S_{\lambda t} = GRU_{S\lambda}(S_{\lambda(t-1)},(C_{\lambda t} \oplus u_{\lambda t} ))
\end{align}

\subsubsection{ Intra-Speaker State}
\label{sssec:subsubhead3}
Naturally, the intra-speaker state is easily observed, felt, and understood by the other participants. More concretely, this state is usually about the expressions, reactions, and responses [3]. Since utterances constantly interfere with each other, we construct an attentive interactive module called \emph{Attention Interactive Dependency}. For the utterance at time $t$ the intra-speaker $I_t$ is updated by the previous intra-speaker state $I_{t-1}$, attentive contextual vector $a_t$, and utterance $u_t$. At time step $t$, the intra-speaker state $I_t$ can be computed as:
\begin{align}
I_{t} = GRU_I(I_{t-1}, (a_{t}\oplus u_{t})).
\end{align}
\subsubsection{Emotion State}
\label{sssec:subsubhead4}
The emotion state performs utterance’s emotion and emotional category. For the utterance at time $t$ the emotion state $E_t$ depends upon the previous emotion state $E_{t-1}$ and the composite of the attentive contextual state $C_t$, self-speaker state $S_t$, intra-speaker $I_t$, and the utterance $u_t$. Then the emotion state $E_t$ can be computed as: 
\begin{align}
E_{t} = GRU_E(E_{t-1}, (C_{t}\oplus S_{t}\oplus u_{t})).
\end{align}


\subsubsection{Classification}
\label{sssec:subsubhead5}
The final output emotion state \emph{E}$_t$ are fed into two fully-connected layers with a residual connection. To train the model, categorical cross-entropy loss with softmax activation in the last layer is used as the loss function, and L2-regularization is applied by adding a penalty in the cost function. 

\section{Experiments}
\label{sec:majhead}

\subsection{Database and Metrics}
\label{ssec:subhead4}

We used the multi-modal and multi-speaker conversational dataset, namely Multi-modal EmotionLines Dataset (MELD)[4]. MELD contains acoustic, textual, and visual information for 13798 utterances and 1433 conversations from the TV series ``Friends''. There are seven emotion categories including: anger, disgust, sadness, joy, neutral, surprise and fear. The dataset is split into the training set, validation set and test set which contains 9989, 1109, and 2610 utterances, respectively [4]. In this work, we only used acoustic modality in related experiments [4]. Due to the natural imbalance across various emotions, we chose a weighted average F1 measure as the evaluation metric.

\begin{table*}[htbp]
	\centering
	{ Table 1: Performance comparison with the state-of-the-art and baselines on MELD}
	\begin{tabular}{c c c c c c c c c}
		\toprule
		\hline
		Method & Anger & Disgust & Fear & Joy & Neural & Sadness & Surprise & w-average F1 (\%) \\
		\hline
		\midrule
		bc-LSTM & 21.9 & 0 & 0 & 0 & 66.1 & 0 & 16 & 36.4 \\
		CMN & 29.6 & 0 & 0 & 11.8 & 67 & 0 & 2.8 & 38.3 \\
		ICON & 31.5	& 0	& 0	& 8.6 & 66.9 & 0 & 0 & 37.7 \\
		DialogueRNN & 32.1 & 5.1 & 0 & 11.2 & 53 & 8.3 & 15.6 & 34 \\
		M2FNet & - & - & - & - & - & - & - & 39.6 \\
		MMTr & - & - & - & - & - & - & - & 38.8 \\ 
		\textbf{Proposed method} & \textbf{32.6} & \textbf{0} & \textbf{0} & \textbf{25.21} & \textbf{63.85} & \textbf{6.28} & \textbf{14.58} & \textbf{40.9} \\
		\hline
		\bottomrule
	\end{tabular}
	\label{tab:table2}
\end{table*}

\subsection{Baselines and State-of-the-Art}
\label{ssec:subhead5}
Totally 6 state-of-the-art methods are compared in the experiments to verify the effectiveness of our proposed approach (Table 1). bc-LSTM are traditional context dependent sentiment analysis [4]. While CMN [13], ICON[16] and DialogueRNN [17] are mainly to model speaker dynamic, M2FNet [18] and MMTr [19] are attention-based method. The brief introductions of these 6 compared methods are presented as follows:

$\bullet$ Bidirectional Contextual LSTM (bc-LSTM) leverages an utterance-level LSTM to learn context dependency. However, the contextual-LSTM model does not accommodate inter-speaker dependencies [4]. 

$\bullet$ Conversational Memory Network (CMN) extracts utterance context from dialogue history information using speaker-dependent gated recurrent units. Such memories are then merged leveraging attention-based hops to capture inter-speaker dependencies [13]. 

$\bullet$ Interactive Conversational Memory Network (ICON) extends CMN to model the self- and inter-speaker sentiment influences and store contextual summaries by using an interactive memory network [16]. 

$\bullet$ DialogueRNN utilizes \emph{GRU} to capture the participant emotional states throughout conversation and the sentence-context representation between speakers [17].

$\bullet$  M2FNet employs a multi head attention-based fusion mechanism to learn emotion-rich latent information of the audio, text and visual modality [18]. 

$\bullet$  MMTr acquires emotional cues at both levels of the speaker’s self-context and contextual context and learns the information interactions between multiple modalities [19].

\subsection{Acoustic Features}
\label{ssec:subhead6}
The audio files were resampled to 16 kHz. In order to extract the Mel-spectrogram, a linear spectrogram was first computed using the Short-Time Fourier Transform with a 25 ms Hamming window and a 10 ms overlapping. After that, a bank of 64 Mel-filters was applied in the frequency range of 125-7500 Hz. These features were then segmented into non-overlapping examples of 0.96 seconds. Especially, the shorter samples were zero-padded b
efore transformation into log Mel-spectrogram. Finally, Each segment was fed into a pre-trained VGGish model obtaining a 128-dimensional embedding vector  for
identifying emotion.

\subsection{Model Configuration}
\label{ssec:subhead7}
We implemented our proposed model using the Pytorch 1.11.0 framework. The model was trained with Adam optimizer with an initial learning rate of 1e-4 and a batch size of 32. Cross-entropy loss was utilized as the loss function. To prevent overfitting, the network was regularized by L2 norm of the model’s parameters with a weight of 3e-4. 

\section{Results and Discussion}
\label{sec:print}

Overall results of our proposed model and the previous state-of-the-art result are presented in Table1. The performance of our method reaches F1 score of 40.9\%. The proposed model is the best performance compared to the state-of-the art and baseline methods and achieves substantial improvements on Anger and Joy (Table 1). Ablation study (see Table 2) verifies the importance of the inter- and intra-speaker state that it is better to consider among attentive context-sensitive dependency, intra- and inter-speaker influence for conversational emotion recognition than only considering context information from single speaker alone. Besides, the model’s performance is worse when the attentive contextual state module is removed, which indicates that modeling of attentive long-term context dependency is more critical than the modeling of intra-speaker-sensitive interactions.

\begin{center} 
{\small Table 2:Albation study on the MELD dataset}
\end{center}
\begin{center}
 \begin{tabular}{l c c c}
 \hline
 Method &  w-average F1  \\ 
 \hline
 \emph{w/o} pretained VGGish  & 36.1  \\
 
\emph{w/o} attentive contextual state & 37.5 \\

\emph{w/o} self-speaker state & 38.9\\

\emph{w/o} intra-speaker state & 38.3 \\

 \hline
\end{tabular}
\end{center}

Quantitatively, our method uses segment-based feature representation for utterance-level classification. Emotions are brief in duration, most lasting only up to a few seconds. Thus, a segment-wise approach is beneficial for capturing temporal information and short-term emotion interaction. In addition, the pre-trained VGGish model retains a major portion of their prior knowledge for better high-level emotional audio features extraction. In particular, a pre-trained model is useful in the current situation where the dataset has limited size and is unbalanced.

Moreover, our model infuses attentive contextual representation from surrounding utterance history, adding it to self-speaker state and intra-speaker influence to capture emotional dynamics on multi-turn conversations. In this regard, it is more advantageous than prior approaches like bc-LSTM, which often loses the ability to determine this kind of situation. For example, character Bob suddenly shifts emotions from neutral to happy when Jill told him ``I am getting married soon'' with a rising tone shade of joy and excited. The attention mechanism is applied to amplify important global contextual conversational dependence (see Fig. 4). In addition, the inter-speaker interactions are either synchronous (for example, cheer after speaking good news) or asynchronous (for example, laughter after speaking something funny). Further analysis on why our method prediction is better could shed light on attentive contextual dependency and speaker-sensitive influence.

\begin{figure}[t!]

\begin{minipage}[b]{1.0\linewidth}
  \centering
  \centerline{\includegraphics[width=8.5cm]{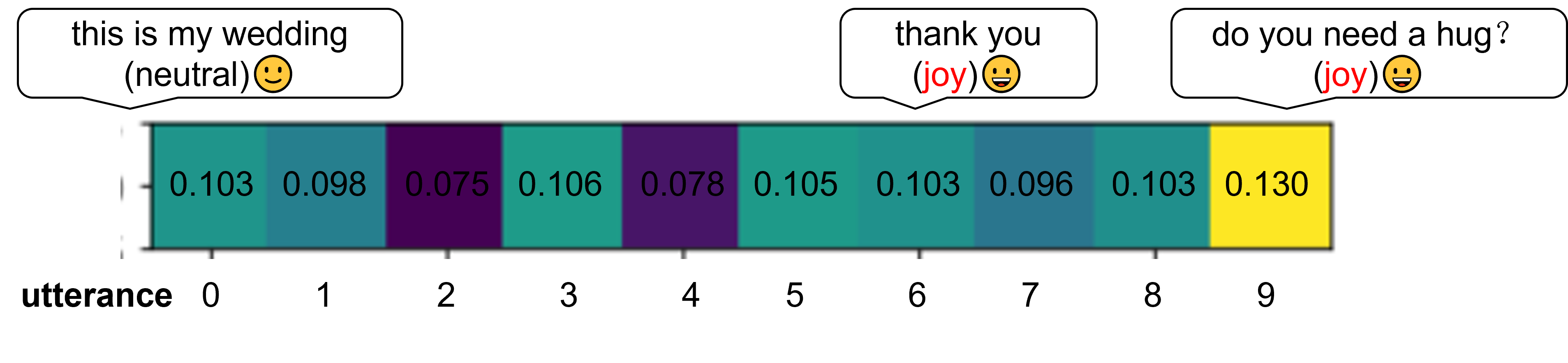}}
\end{minipage}
\caption{\small Attention weight visualization of our model from the cases in MELD.}
\end{figure}

\section{Conclusion}
\label{sec:page}

In this paper, we proposed the aggregation of segment-level speech spectrogram for utterance-level emotion classification in conversations. It capitalized on inferring the contextual information that incorporates dynamic self-, intra- and inter-speaker influence. An attention-based mechanism was employed to determine the important contextual-sensitive information from surrounding utterances history. The bi-directional \emph{GRU} was used to capture contextual dependency, self- and inter-speaker influence. The experiment results demonstrate the effectiveness of the proposed model. 

There are some limitations to our method. The generalization, detailed ablation studies and analysis in several other benchmark datasets need to be explored further to increase dataset’s scope and advance research in ERC. Additionally, it is known that text in speech, facial expressions, body movements, and other modalites all convey emotions, and hence can provide additional information. A promising further direction of research would be to apply recently advanced approaches to transfer the knowledge from multimodal systems to unimodal systems, thus improving the performance for unimodal systems.

\end{document}